\pdfoutput=1
\documentclass{article}
\usepackage{spconf,amsmath,epsfig}
\usepackage[ruled]{algorithm2e}
\usepackage{verbatim}
\usepackage{booktabs}
\usepackage{bm}
\usepackage{graphicx}
\usepackage{amssymb}
\usepackage[T1]{fontenc}
\usepackage{multirow}
\usepackage{makecell}
\usepackage{url}
\usepackage[pagebackref,breaklinks,colorlinks]{hyperref}
\newcommand{\tabincell}[2]{\begin{tabular}{@{}#1@{}}#2\end{tabular}}

\let\OLDthebibliography\thebibliography
\renewcommand\thebibliography[1]{
  \OLDthebibliography{#1}
  \setlength{\parskip}{0pt}
  \setlength{\itemsep}{0pt plus 0.3ex}
}

\begin{document}\sloppy

\def\x{{\mathbf x}}
\def\L{{\cal L}}

\title{Accelerating Diffusion Sampling with Classifier-based Feature Distillation}
%
\name{Wujie Sun, Defang Chen, Can Wang, Deshi Ye, Yan Feng, Chun Chen}
\address{Zhejiang University, Hangzhou, China\\
sunwujie@zju.edu.cn}

\maketitle

\begin{abstract}
Although diffusion model has shown great potential for generating higher quality images than GANs, slow sampling speed hinders its wide application in practice. Progressive distillation is thus proposed for fast sampling by progressively aligning output images of $N$-step teacher sampler with $N/2$-step student sampler. 
In this paper, we argue that this distillation-based accelerating method can be further improved, especially for few-step samplers, with our proposed \textbf{C}lassifier-based \textbf{F}eature \textbf{D}istillation (CFD). Instead of aligning output images, we distill teacher's sharpened feature distribution into the student with a dataset-independent classifier, making the student focus on those important features to improve performance. We also introduce a dataset-oriented loss to further optimize the model. Experiments on CIFAR-10 show the superiority of our method in achieving high quality and fast sampling. Code is provided at \url{https://github.com/zju-SWJ/RCFD}.
\end{abstract}
\begin{keywords}
Diffusion model, knowledge distillation, image generation, fast sampling
\end{keywords}

\section{Introduction}
Image generation is an important research field in computer vision and various models have been invented, such as generative adversarial networks (GANs)~\cite{saxena2021generative} and diffusion models~\cite{cao2022survey}. 
The adversarial nature of GANs requires careful architecture and hyper-parameter selection to stabilize the model training, while the recent diffusion models can overcome these weaknesses and achieve better performance~\cite{dhariwal2021diffusion}. However, diffusion models require a greatly slower iterative sampling to get the final denoised images. How to accelerate the sampling efficiency becomes a critical issue.

Two main acceleration directions are training-free sampling and training scheme~\cite{cao2022survey}. Training-free sampling~\cite{song2020denoising,liu2021pseudo,lu2022dpm} aims to propose efficient sampling methods to boost sampling speed for the pre-trained diffusion models, while training scheme methods~\cite{salimans2021progressive,nichol2021improved,kong2021fast} require additional training, but it gives model the potential for more powerful performance. 

Recently, knowledge distillation-based training scheme methods~\cite{salimans2021progressive,luhman2021knowledge} have exhibited strong capabilities in fast sampling and high performance, surpass other methods~\cite{song2020denoising,liu2021pseudo,lu2022dpm,nichol2021improved,kong2021fast} with large margins. Inspired by the idea of distilling the knowledge in a powerful teacher model into a compact student model~\cite{hinton2015distilling,chen2022simkd}, Progressive Distillation (PD)~\cite{salimans2021progressive} lets the student sampler mimic the teacher sampler's two-step output with a single step. In this way, the sampler maintains a decent performance when progressively halving its sampling steps. However, little work has been done upon this. 

\begin{figure*}[t]
    \centering
    \includegraphics[width=\linewidth]{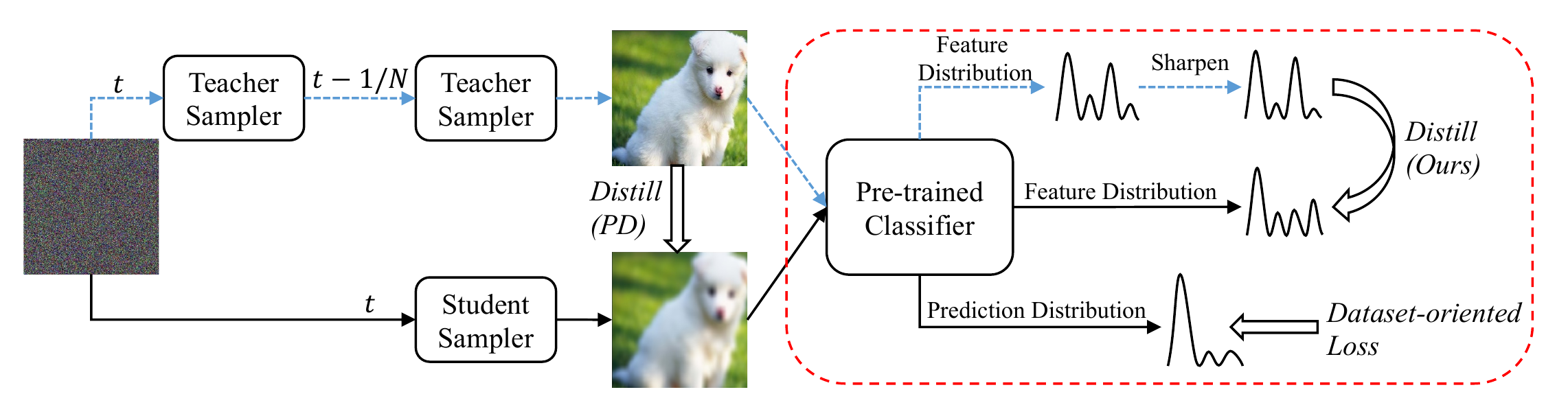}
    \caption{Overall framework of our method. Instead of directly aligning images as PD, we use the pre-trained classifier to get feature and prediction distributions. The teacher's feature distribution is sharpened and distilled to the student, which involves no category information, giving our method wide applicability. The prediction distribution is guided by dataset-oriented loss to further improve performance, which can be used when a pre-trained classifier on the target dataset is available.}
    \label{fig:framework}
\end{figure*}

In this paper, by using an additional classifier, we further demonstrate the power of knowledge distillation in speeding up diffusion sampling. 
We argue that strictly aligning the individual pixels in output images of the student and teacher samplers is difficult, especially for student samplers with few sampling steps. 
With the help of a classifier, we can get the high-level feature distributions based on the images output by teacher and student. By calculating the KL-divergence of these two distributions, student is able to focus on those important features (which are closely related to image composition), thus reducing the learning burden and ensuring the consistency of the image. We name it \textbf{C}lassifier-based \textbf{F}eature \textbf{D}istillation (CFD). Notice that at this point, our classifier is NOT necessarily trained on the target dataset, since it is only used for feature extraction and does not involve category information. This allows our method to be applied to datasets that are not used for classification. Such classifier, which does not require adversarial training and pre-training on the target dataset, makes our work very different from previous works with classifiers for image generation and refinement~\cite{santurkar2019image,ganz2021bigroc}. For classifiers trained on the target dataset, we further propose \textbf{R}egularized \textbf{CFD} (RCFD) which combines CFD with entropy and diversity losses to further optimize model performance. We provide an overview of our method in Figure~\ref{fig:framework}.

Our contributions can be summarized as follows:
\begin{itemize}
    \item We propose a novel classifier-based distillation method for speeding up diffusion sampling speed.
    \item Our method does not involve adversarial training, and does not require the classifier to be pre-trained on the target dataset, making our method easy to use and widely applicable.
    \item Experiments on CIFAR-10 show that our method outperforms SOTA methods with large margins.
\end{itemize}

\section{Related work}
\subsection{Diffusion model}
Diffusion models aim to sample high-quality images from random noises, which contains two processes: training and sampling. A standard training process is proposed by DDPM~\cite{ho2020denoising}. The well-trained network with parameter $\theta$ could take noisy image $\textbf{z}_t$ and time $0\leq t\leq 1$ as inputs, and outputs the predicted denoised image $\textbf{x}_t=\theta(\textbf{z}_t,t)=\theta(\textbf{z}_t)$. Starting from $t=1$, the sampling process is then repeated $N$ times to get the final generated image. Since such sampling process is very time-consuming, DDIM~\cite{song2020denoising} proposes an implicit sampling to speed up, which can be represented as
\begin{equation}\label{equ:ddim}
    \textbf{z}_s=\alpha_s  \underbrace{\theta(\textbf{z}_t)}_{\text{predicted denoised image $\textbf{x}_t$}} +\sigma_s \underbrace{\frac{\textbf{z}_t-\alpha_t \theta(\textbf{z}_t)}{\sigma_t}}_{\text{direction pointing to $\textbf{z}_t$}},
\end{equation}
where $\alpha$ and $\sigma$ are pre-defined time-related functions, $\textbf{z}_0$ is the final denoised image, and $0\leq s<t\leq 1$. We provide a more detailed explanation in Appendix~\ref{appendix:diffusion}. Based on DDIM, Progressive Distillation (PD)~\cite{salimans2021progressive} uses knowledge distillation to improve sampling speed. Other methods such as PNDMs~\cite{liu2021pseudo} and DPM-Solver~\cite{lu2022dpm} also manage to speed up sampling, but fail to outperform PD with huge margins.

\subsection{Knowledge distillation}
Knowledge distillation~\cite{hinton2015distilling} is an efficient method for model compression. Diverse knowledge such as logits~\cite{hinton2015distilling} and intermediate features~\cite{chen2021cross}, can be transferred from a superior teacher model to a compact student model. In addition, online knowledge distillation~\cite{chen2020online} introduces multiple training models, while self-distillation~\cite{furlanello2018born} contains only a single model architecture. Although knowledge distillation has a wide applications such as image classification~\cite{hinton2015distilling} and semantic segmentation~\cite{shu2021channel}, distillation for fast diffusion sampling~\cite{salimans2021progressive} has rarely been explored yet. We believe this field holds great promise.

\subsection{Classifier for image generation}
Classifier is important for image classification. Recent works show that it can also be applied to image generation~\cite{santurkar2019image,ganz2021bigroc}. However, these methods need a robust classifier with adversarial training, which increases training difficulty. Classifier is also used in diffusion models to provide class-related guidance and improve performance~\cite{dhariwal2021diffusion}. Different from the above works, in this paper, we use the classifier to extract the feature/prediction distribution of images and transfer it to the student model as knowledge. Such classifier does not require adversarial training and can be pre-trained on a different dataset.

\section{Methodology}
\subsection{Progressive distillation}
Progressive Distillation (PD)~\cite{salimans2021progressive} introduces knowledge distillation to speed up sampling. 
Once teacher sampler with $N$ steps is given, student sampler with $N/2$ steps is trained to speed up sampling. Assuming that the sampling time is now $t$, we can get the predicted denoised image $\textbf{x}^\text{T}$ at time $t-2/N$ by sampling the teacher model for two steps. The detailed derivation for $\textbf{x}^\text{T}$ is provided in Appendix~\ref{appendix:derivation}. The training loss for PD is represented as
\begin{equation}
    L_{\text{PD}}=w_t||\textbf{x}^\text{T}-\theta(\textbf{z}_t)||^2_2,
\end{equation}
where $w_t=\max(\alpha^2_t/\sigma^2_t,1)$ is used for better distillation. 

Directly aligning images is very effective when the sampler has many steps, but it degrades rapidly when there are few steps. We believe that when the sampling steps are small, it becomes difficult for the student to strictly align the pixels on the image, which hinders the model learning. Therefore, we argue that in this situation, the student model should pay more attention to learning the key features associated with images, so as to improve the learning efficiency and quality.

\subsection{Classifier-based feature distillation}
A classifier $cls$ is usually composed of two parts, feature extractor $extr$ and fully connected layers. 

Instead of aligning $\textbf{x}^\text{T}$ and $\theta(\textbf{z}_t)$ as PD~\cite{salimans2021progressive}, we use a classifier to extract features and use them as transferred knowledge. To be more specific, student's output image $\textbf{x}^\text{S}=\theta(\textbf{z}_t)$ and teacher's output image $\textbf{x}^\text{T}$ are input to the same extractor $extr$, and output the last features before the fully connected layers, which can be represented as
\begin{equation}
\begin{aligned}
    \textbf{F}^{\text{S}}=extr(\textbf{x}^\text{S}),\quad
    \textbf{F}^{\text{T}}=extr(\textbf{x}^\text{T}).
\end{aligned}
\end{equation}

After that, we convert feature into distribution using softmax function $\sigma(\cdot)$, and calculate the KL-divergence between teacher and student feature distributions
\begin{equation}\label{equ:kl}
\begin{aligned}
    L_\text{CFD}=\text{KL}\left(\sigma(\textbf{F}^\text{S}),\sigma^\tau(\textbf{F}^\text{T})\right),
\end{aligned}
\end{equation}
where temperature $0<\tau\leq 1$ is used to sharpen the distribution. Note that $\tau$ is only applied to teacher feature distribution, which we find to be more effective than applying to both distributions. In this case, the upper limit of student performance is no longer the teacher, so in some cases (see section \ref{exp:result}), students can even surpass the teacher model!

Different from the L2 distance used in PD, KL divergence can give large feature values greater weight in gradient descent, thus helping the model focus more on aligning these features. 
After the image is input into the feature extractor, the features change from the shallow fine-grained features to the deep coarse-grained features as the layer increases. Deep features contain more semantic information related to categories, which is crucial for image composition. 
By aligning important teacher features, and reducing the interference of irrelevant features on model training, students with poor ability can learn more useful knowledge to generate high-quality images and improve performance. 

Note that the loss in Equ.~\ref{equ:kl} is NOT oriented to a specified dataset, since we only use the feature extractor  and do not include the subsequent fully connected layers for classification. This advantage makes our proposed distillation method can be extended to more datasets, such as CelebA and LSUN bedrooms. Next, we further introduce dataset-oriented loss to help the model better improve performance.

\noindent \textbf{Dataset-oriented loss.} 
For a $N$-step sampler, as the sampling step increases, the image obtained by $\theta(\textbf{z}_t)$ tends to be clearer. A clearer denoised image in the early steps will benefit the subsequent sampling steps.

\begin{figure}[t]
    \centering
    \includegraphics[width=\linewidth]{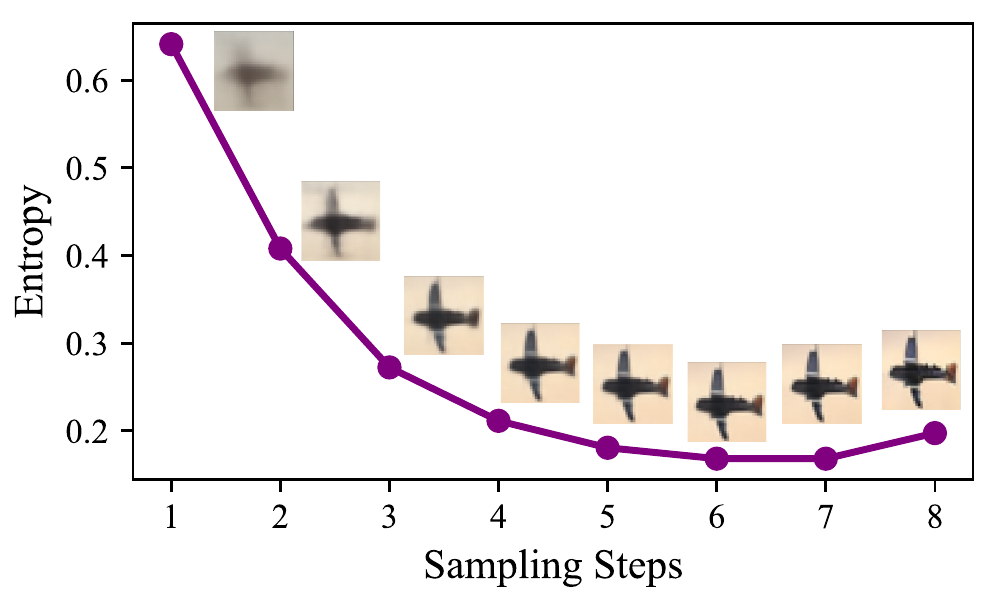}
    \caption{
    Entropy evaluation on CIFAR-10 using 8-step sampler with different sampling steps. The classifier is ResNet18 and the entropy is calculated by averaging 4096 generated images.
    We give an illustration of the denoised image from the same noise initialization.
    }
    \label{fig:entropy}
\end{figure}
By feeding the images obtained from each sampling step into a classifier, we can calculate the entropy as follows
\begin{equation}
    L_{\text{entropy}}=-\sum^{C}_{c=1}\textbf{p}_c\log\textbf{p}_c, \quad
    \textbf{p}=\sigma(cls(\textbf{x}^\text{S})).
\end{equation}
where $C$ is the total number of classes, $\textbf{p}$ denotes prediction results. 
Figure~\ref{fig:entropy} shows that
sampling with fewer steps yield a larger entropy and generate more blurred images.
This means if we minimize the entropy of prediction results, we could get relatively clearer images, especially for early sampling steps.

In addition, with the progressive distillation, it inevitably makes the current sampler's output image distribution deviate more and more from the original optimal one. Since the dataset we used is balanced, we expect the predicted probability to remain equal for each class within each batch:
\begin{equation}
\begin{aligned}
    L_{\text{diversity}}=\sum^C_{c=1}\hat{\textbf{p}}_c\log\hat{\textbf{p}}_c,\quad
    \hat{\textbf{p}}=\frac{\sum^B_{b=1}\textbf{p}}{B},
\end{aligned}
\end{equation}
where $B$ is the batch size. These two losses do not involve teacher guidance and are thus less effective when used alone. But better results can be achieved by combining them with $L_\text{CFD}$.

\noindent \textbf{Overall loss.} The overall loss function can be represented as
\begin{equation}
    L_\text{RCFD}=L_\text{CFD}+\beta[\gamma L_\text{entropy}+(1-\gamma)L_\text{diversity}],
\end{equation}
where $\beta$ and $\gamma$ are hyper-parameters, and RCFD stands for \textbf{R}egularized \textbf{C}lassifier-based \textbf{F}eature \textbf{D}istillation.

\section{Experiment}
\subsection{Setting}
In this section, we use CIFAR-10 to validate the superiority of our method. We use the cosine schedule introduced in~\cite{nichol2021improved} to calculate $\alpha_t$ and $\sigma_t$. We use the U-Net~\cite{ronneberger2015u} as the diffusion model. ResNet18~\cite{he2016deep} and DenseNet201~\cite{huang2017densely} are used as the classifiers. The base diffusion model is trained with 1024 steps. More details are provided in Appendix \ref{appendix:experiment}.

We compare our method with DDIM~\cite{song2020denoising}, Progressive Distillation (PD)~\cite{salimans2021progressive}, PNDMs~\cite{liu2021pseudo}, and DPM-Solver~\cite{lu2022dpm}. 
The distillation-based acceleration method requires iterative training to halve sampling steps. Based on results in~\cite{salimans2021progressive} and our own experiments, we find that performance changes rapidly in distillation from 8-step to 1-step. So we focus on distillation process starting from 8-step and train the teacher model as PD~\cite{salimans2021progressive} from 1024 to 8 steps without the classifier. We re-implemented DDIM and PD for better comparison. Note that the PD performance we reported in Table~\ref{tab:result} is different from the original paper~\cite{salimans2021progressive} because we failed to train a good base model on the U-Net architecture used by PD. Therefore, we chose the architecture introduced in DDPM~\cite{ho2020denoising}, and used a smaller distillation iterations to reduce training overhead.

\begin{table}[t]
    \centering
    \begin{tabular}{c|c|c|c}
        \hline
        \tabincell{c}{Sampling \\Steps} & Method & IS $\uparrow$ & FID $\downarrow$ \\
        \hline
        \multirow{3}*{1} & \textbf{RCFD-DenseNet201} & \textbf{8.87} & \textbf{8.92} \\
        ~ & RCFD-ResNet18 & 8.56 & 12.03 \\
        ~ & PD (ICLR 2022) & 7.88 & 15.06 \\
        \hline
        \multirow{3}*{2} & \textbf{RCFD-DenseNet201} & \textbf{9.19} & \textbf{5.07} \\
        ~ & RCFD-ResNet18 & 9.09 & 6.12 \\
        ~ & PD (ICLR 2022) & 8.70 & 7.42 \\
        \hline
        \multirow{3}*{4} & \textbf{RCFD-DenseNet201} & \textbf{9.34} & \textbf{3.80} \\
        ~ & RCFD-ResNet18 & 9.24 & 4.24 \\
        ~ & PD (ICLR 2022) & 9.04 & 4.83 \\
        \hline
        \multirow{2}*{8} & PD (ICLR 2022) & 9.14 & 4.14 \\
        ~ & DDIM (ICLR 2021) & 8.14 & 20.97 \\
        \hline
        10 & PNDMs (ICLR 2022) & - & 7.05 \\
        \hline
        12 & \tabincell{c}{DPM-Solver (NIPS 2022)} & - & 4.65 \\
        \hline
        1024 & DDIM (ICLR 2021) & 9.21 & 3.78 \\
        \hline
    \end{tabular}
    \caption{Performance comparison with state-of-the-art methods on CIFAR-10. Higher IS and lower FID are better. Results are the average of 3 runs.}
    \label{tab:result}
\end{table}

\subsection{Result}
\label{exp:result}
The result is shown in Table~\ref{tab:result}. As we can see, distillation-based methods (RCFD and PD) surpass other methods with large margin (4-step distillation-based samplers can achieve the performance of other samplers with 10+ steps). Also, the difference between the 8-step sampler obtained by PD and the 1024-step DDIM sampler (base diffusion model) is small, indicating the effectiveness of distillation. 

In addition, RCFD with DenseNet201 achieved 6.14 ($\downarrow$40.7\%), 2.35 ($\downarrow$31.6\%), and 1.03 ($\downarrow$21.3\%) FID improvement compared to PD in the 1, 2, and 4-step samplers, respectively, demonstrating its superiority. 
Also, with the help of the classifier, we offer the possibility for the student sampler (4-step of RCFD-DenseNet201) to significantly outperform its teacher (8-step sampler obtained from PD).

\subsection{Ablation study}
In this section, we perform some ablation studies to verify the importance of each component in our method. If not specified, we use ResNet18 as the classifier and use the 8-step sampler trained by PD as the teacher to train a 4-step student.

\begin{table}[t]
    \centering
    \begin{tabular}{c|ccc|c|c}
        \hline
        Method & $L_\text{CFD}$ & $L_\text{entropy}$ & $L_\text{diversity}$ & IS $\uparrow$ & FID $\downarrow$ \\
        \hline
        \multirow{5}*{RCFD} & \checkmark &  &  & 9.14 & 4.42 \\

        ~ & & \checkmark &  & 2.18 & 330.27 \\

        ~ & &  & \checkmark & 1.22 & 308.61 \\

        ~ & & \checkmark & \checkmark & 5.87 & 92.07 \\

        ~ & \checkmark & \checkmark & \checkmark & \textbf{9.24} & \textbf{4.24} \\
        \hline
        PD & & & & 9.04 & 4.83 \\
        \hline
    \end{tabular}
    \caption{Impact of each loss on performance.}
    \label{tab:component}
\end{table}

\begin{figure}[t]
    \centering
    \includegraphics[width=\linewidth]{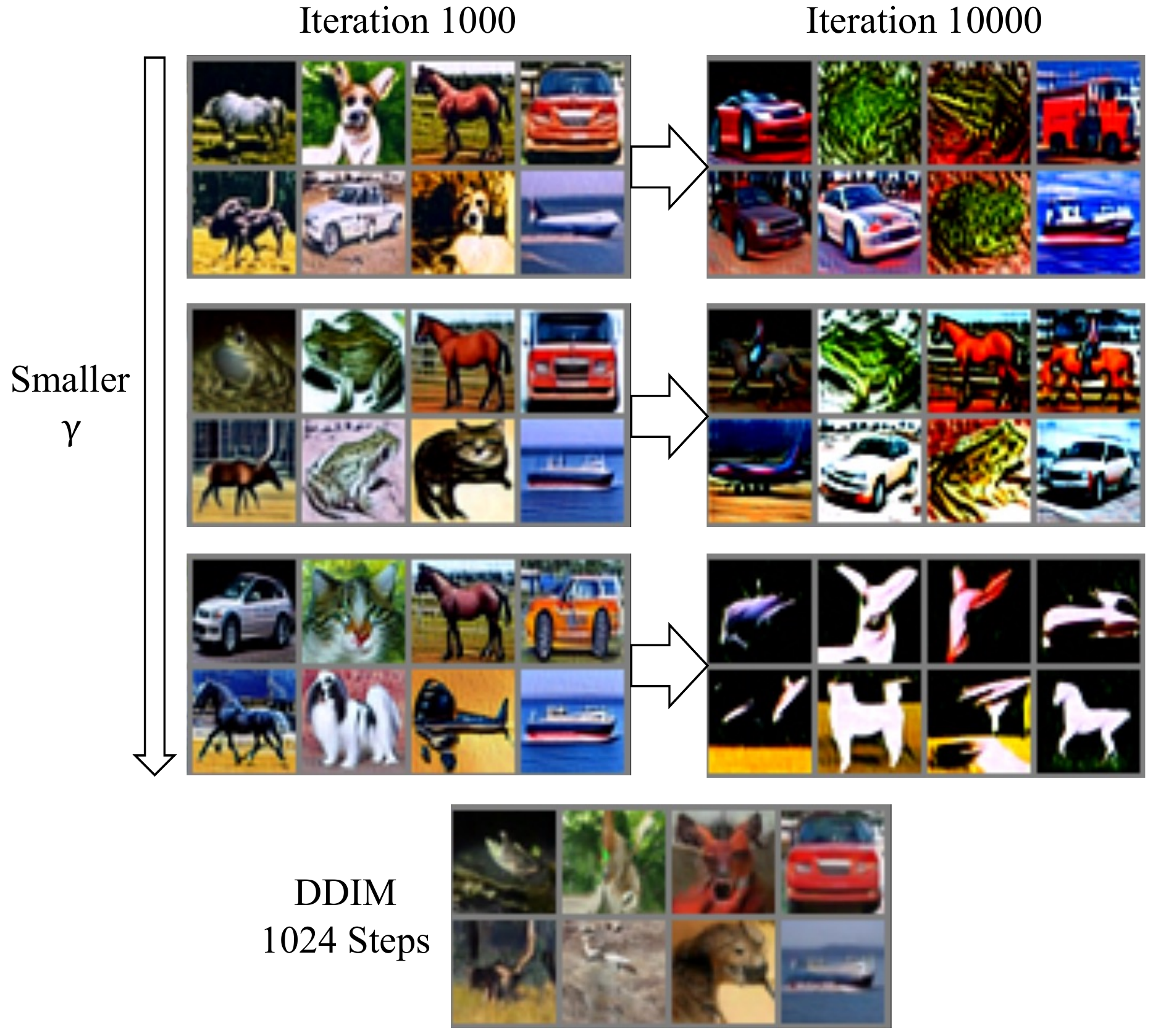}
    \caption{The first three rows indicate the images of different training stages obtained by only using different scales of $L_\text{entropy}$ and $L_\text{diversity}$. 
    It can be seen that in the absence of $L_\text{CFD}$, the visual quality of the images at early training stage surpasses even the sampler with 1024 steps of DDIM, but those images are easily collapsed latter.}
    \label{fig:compare}
\end{figure}

\noindent \textbf{Ablation study on each loss.} Three losses are included in our method, $L_\text{CFD}$, $L_\text{entropy}$, and $L_\text{diversity}$. $L_\text{CFD}$ is a dataset-independent loss, which introduces classifier-based distillation to align student's feature distribution with teacher's sharpened feature distribution. The latter two are dataset-oriented losses, where $L_\text{entropy}$ is used to generate clearer images and $L_\text{diversity}$ maintains the class balance. 

As we can see from Table~\ref{tab:component}, with only $L_\text{CFD}$, we can already achieve better performance than PD. Although good results cannot be achieved using $L_\text{entropy}$ and $L_\text{diversity}$ when $L_\text{CFD}$ is not available, optimal performance can be achieved by combining all three terms. The reason is that the teacher constraint ($L_\text{CFD}$) will prevent the generated images from being too abstract and meaningless during training (as shown in Figure~\ref{fig:compare}), which improves the model performance.

\noindent \textbf{Ablation study on classifier.} 
In this section, we try different classifiers and see how the performance changes. As shown in Table~\ref{tab:cls}, no matter what classifier we use, we can achieve better results than PD. However, if possible, it is better to train the classifier on the target image generation dataset. In addition, as the classifiers become more and more powerful, they also help the student samplers produce higher quality images, which even achieve significantly better FID than the teacher. We believe that for a more powerful classifier, it will extract more accurate and meaningful features, therefore, it provides students with more effective knowledge for distillation, thus helping students produce better images.

\begin{table}[t]
    \centering
    \begin{tabular}{c|c|c|c}
        \hline
        Pre-trained Dataset & Classifier & IS $\uparrow$ & FID $\downarrow$ \\
        \hline
        \multirow{3}*{CIFAR-10} & ResNet18  & 9.14 & 4.42 \\
        ~ & ResNet50 & 9.16 & 4.24 \\
        ~ & DenseNet201 & \textbf{9.34} & \textbf{3.80} \\
        \hline
        ImageNet & ResNet50 & 9.05 & 4.62 \\
        \hline
    \end{tabular}
    \caption{Impact of different pre-trained classifiers on CIFAR-10 image generation performance. For better comparison, we use $L_\text{CFD}$ only.}
    \label{tab:cls}
\end{table}

\noindent \textbf{Ablation study on $L_\text{CFD}$.} 
In our method, we align feature distributions rather than prediction distributions since the former is dataset-independent and achieve better results. 
\begin{figure}[t]
    \centering
    \includegraphics[width=\linewidth]{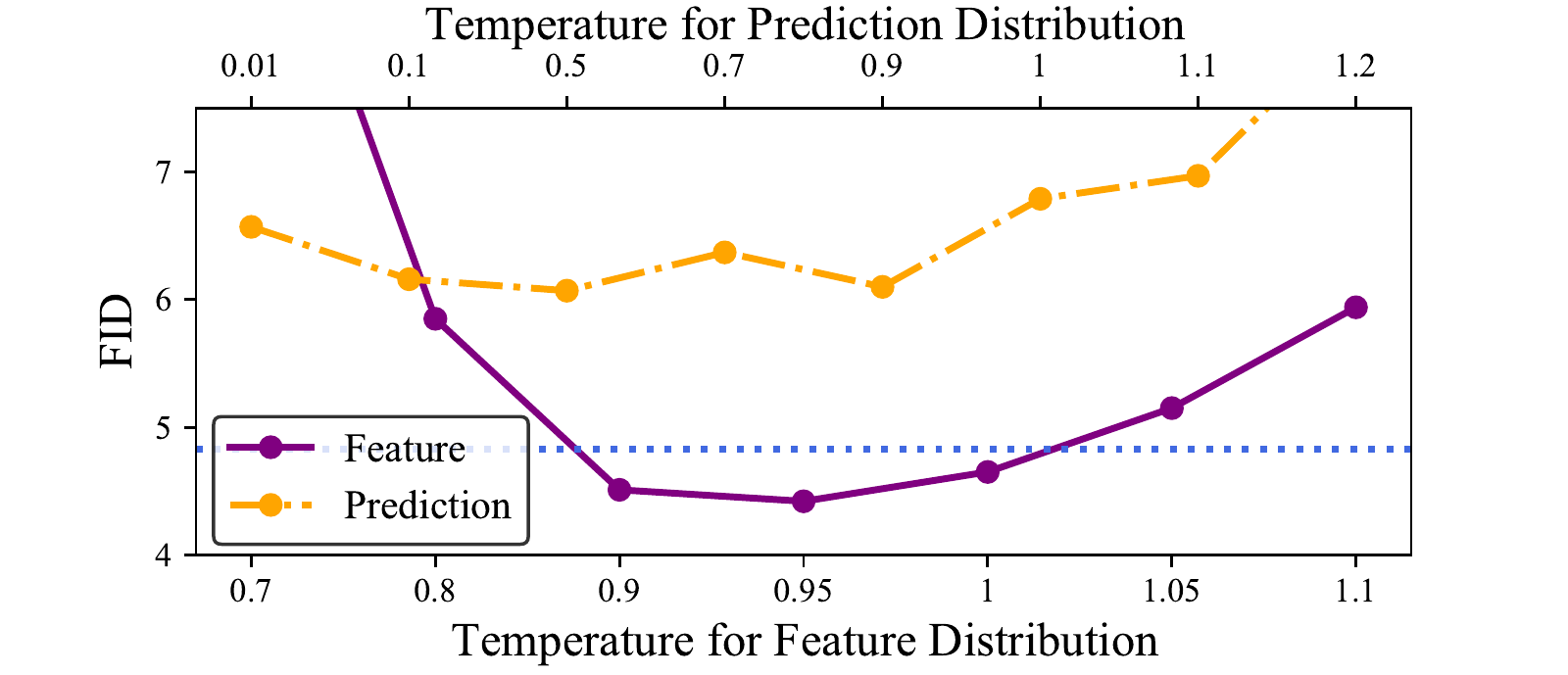}
    \caption{Impact of softmax temperature on performance. The FID of PD is shown by the blue dotted line. For better comparison, we use $L_\text{CFD}$ only.}
    \label{fig:single}
\end{figure}

In this section, we provide the performance comparison of these two approaches under different softmax temperatures.
Figure~\ref{fig:single}
shows that aligning student's feature distribution with teacher's slightly sharpened feature distribution (temperature $0.9 \leq \tau\leq 1$) obtain better results than PD and almost always outperform distilling the prediction distributions.
For the feature distribution, over large temperature will make the teacher's feature distribution tend to be uniformly distributed, hindering the learning of important features and making the image meaningless, while over small temperature makes few features to be highlighted, making the image too abstract and causing performance degradation. 
For the prediction distribution, since it has smaller constraints compared to the feature distribution (i.e., different feature distributions may yield the same prediction results), the learning of image details can be weakened, which leads to bad performance.
\begin{figure}[t]
    \centering
    \includegraphics[width=\linewidth]{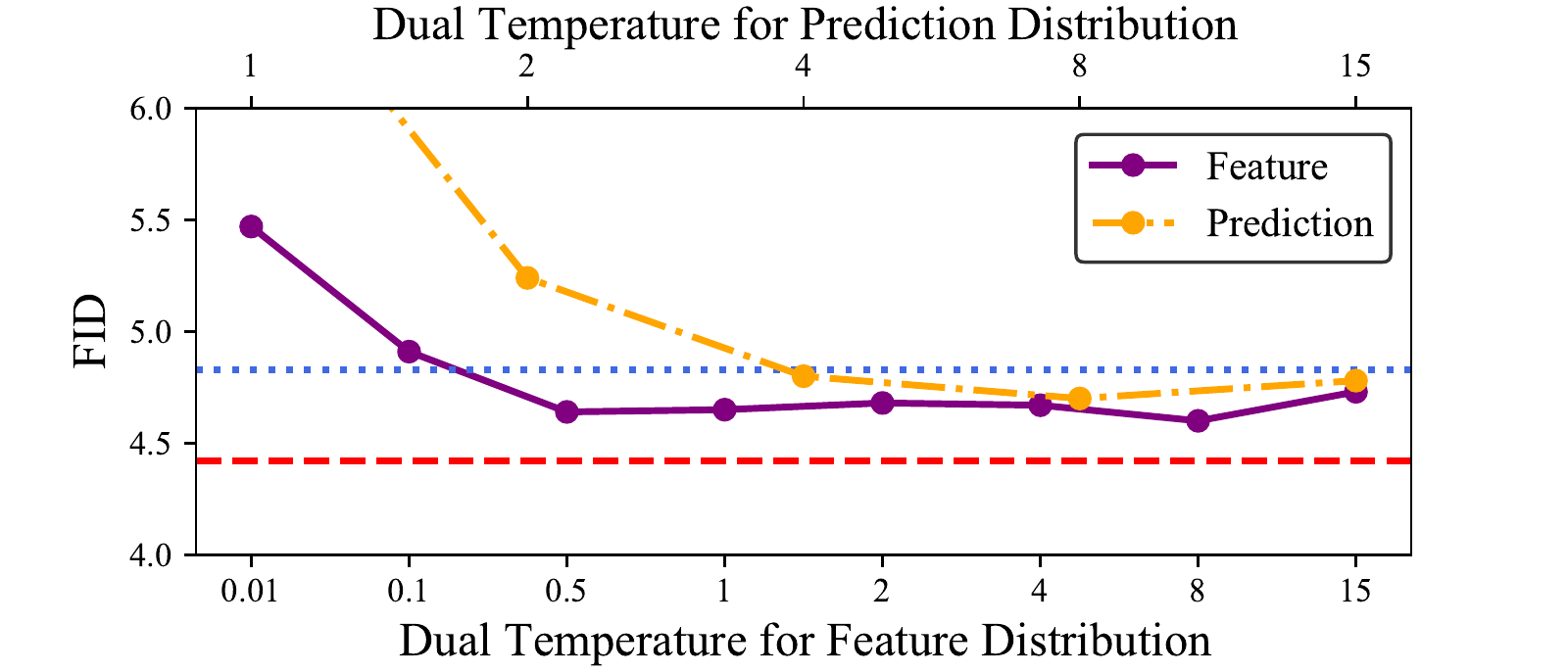}
    \caption{Impact of dual softmax temperature. The FIDs of PD and CFD are shown by the blue dotted line and red dashed line, respectively. For better comparison, we use $L_\text{CFD\_dual}$ only.}
    \label{fig:dual}
\end{figure}

\noindent \textbf{Ablation study on dual softmax temperature.} 
In our method, softmax temperature $\tau$ is only used for the teacher, as shown in Equ. \ref{equ:kl}.
We now apply the same temperatures $\tau$ to both the student and the teacher, and change the loss as
\begin{equation}\label{equ:dual}
\begin{aligned}
    L_\text{CFD\_dual}=\tau^2 \text{KL}\left(\sigma^\tau(\textbf{F}^\text{S}),\sigma^\tau(\textbf{F}^\text{T})\right).
\end{aligned}
\end{equation}

Figure~\ref{fig:dual} show that, for a wide range of temperatures, aligning feature distributions and prediction distributions achieve better performance than PD, but fails to outperform the original $L_\text{CFD}$ which only uses temperature for the teacher. Although large dual temperature helps to improve performance when prediction distributions are aligned, we believe that such aligning (no matter it is feature or prediction distribution) determines that the upper limit of the student is the teacher (unlike traditional knowledge distillation for image classification, there is no additional guidance such as labels during distillation), which limits performance improvement.

\section{Conclusion}
In this paper, we propose a novel classifier-based distillation method to speed up the sampling of the diffusion models. We let student align its feature distribution with teacher's sharpened feature distribution, rather than aligning the generated images. In this way, student can focus on learning important features that make up an image, resulting in even better performance than the teacher. 
This distillation method is also applicable when the classifier is pre-trained on other datasets.
When the classifier pre-trained on the target dataset is available, we propose a dataset-oriented loss to further improve performance. Experiments on CIFAR-10 show the superiority of our method.

\bibliographystyle{IEEEbib}
\bibliography{icme2023template}

\appendix
\begin{figure*}[t]
    \centering
    \includegraphics[width=\linewidth]{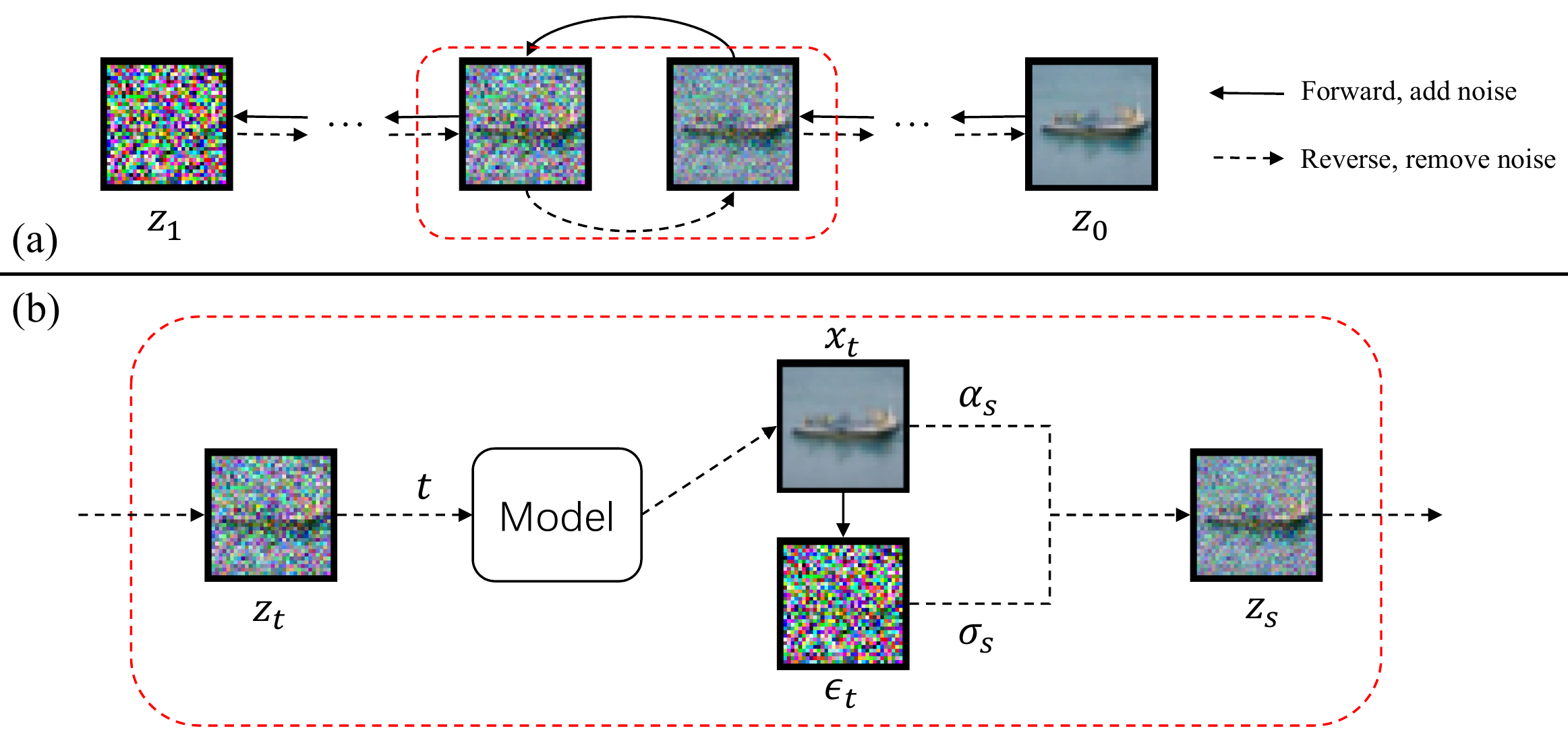}
    \caption{An overview of the training and sampling of diffusion models. (a) The diffusion models contain two processes, forward (turn the image into noise) and reverse (remove noise from the image). Our target is to model the reverse process using the neural network (which can be achieved by using DDPM~\cite{ho2020denoising}), so that we can get images from random noises. (b) Once the model is well trained, we can use it multiple times to get the denoised image. For sampling time $t$, we can get $\textbf{z}_t$ from previous step, and we input $\textbf{z}_t$ and $t$ into the model, which outputs the predicted denoised image $\textbf{x}_t$ (or predicted noise $\epsilon_t$, depending on the training target of the model). Based on the $\textbf{x}_t$ (or $\epsilon_t$), we can get the corresponding $\epsilon_t$ (or $\textbf{x}_t$). After that, we use DDIM (Equ.~\ref{equ:ddim}) to calculate the noisy image $\textbf{z}_s$, which is the model input for next step.}
    \label{fig:diffusion}
\end{figure*}
\section{Training and Sampling of Diffusion Models}~\label{appendix:diffusion}
We provide an overview of the training and sampling of diffusion models in Figure~\ref{fig:diffusion}.

\section{Derivating Denoised Image for Distillation}\label{appendix:derivation}
Assume we have $N$-step teacher, and the current time is $t$, then we can get $t'=t-1/N$ and $t''=t-2/N$. $\textbf{z}_t'$ and $\textbf{z}_t''$ are calculated as
\begin{equation}
    \textbf{z}_{t'}=\alpha_{t'}\eta(\textbf{z}_{t})+\sigma_{t'}\frac{(\textbf{z}_{t}-\alpha_{t}\eta(\textbf{z}_{t}))}{\sigma_{t}},
\end{equation}
\begin{equation}
    \textbf{z}_{t''}=\alpha_{t''}\eta(\textbf{z}_{t'})+\sigma_{t''}\frac{(\textbf{z}_{t'}-\alpha_{t'}\eta(\textbf{z}_{t'}))}{\sigma_{t'}},
\end{equation}
where $\eta$ is the teacher model.

Assume student has denoised image $\textbf{x}^\text{S}$ and gets noisy image $\tilde{\textbf{z}}_{t''}$ in one step. If well aligned, we should have
\begin{equation}
    \textbf{z}_{t''}=\tilde{\textbf{z}}_{t''}=\alpha_{t''}\textbf{x}^\text{S}+\sigma_{t''}\frac{(\textbf{z}_{t}-\alpha_{t}\textbf{x}^\text{S})}{\sigma_{t}}.
\end{equation}

The distillation target $\textbf{x}^\text{T}$ can thus be represented as
\begin{equation}
    \textbf{x}^\text{T}=\textbf{x}^\text{S}=\frac{\textbf{z}_{t''}-(\sigma_{t''}/\sigma_{t})\textbf{z}_{t}}{\alpha_{t''}-(\sigma_{t''}/\sigma_{t})\alpha_{t}}.
\end{equation}

\section{Experiment Details}\label{appendix:experiment}
\subsection{Model Architecture}
The U-Net includes four feature map resolutions (32 $\times$ 32 to 4 $\times$ 4), and it has two convolutional residual blocks per resolution level and self-attention blocks at 8 $\times$ 8 resolution. Diffusion time $t$ is embedded into each residual block. Initial channel number is 128 and is multiplied by 2 at last three resolutions.

\subsection{Performance Evaluation}
We report the Inception Score (IS)~\cite{salimans2016improved} and Fréchet Inception Distance (FID)~\cite{heusel2017gans} results of each method. IS measures the class balance and confidence of the generated images, while FID measures the difference in feature distribution between the generated and real images. Therefore, higher IS and lower FID represent better generated images.

\subsection{Training Setting}
Learning rate (warmup for 5000 iterations) 0.0002, dropout 0.1, batch size 128, ema decay 0.9999, gradient clip 1, total iterations 800000. 

\subsection{Distillation Setting}
\noindent\textbf{Common setting.} Learning rate (cosine annealing) 5e-5, batch size 128, gradient clip 1, total iterations 10000 for 1024 to 4-step and 20000 for 4 to 1-step. 

\noindent\textbf{RCFD (ResNet18) setting.} 
\begin{itemize}
    \item 8 to 4-step: $\tau=0.95, \beta=0.003, \gamma=0.75$.
    \item 4 to 2-step: $\tau=0.95, \beta=0.003, \gamma=0.75$.
    \item 2 to 1-step: $\tau=0.85, \beta=0.003, \gamma=0.5$.
\end{itemize}

\noindent\textbf{RCFD (DenseNet201) setting.} Since introducing dataset-oriented loss makes it more difficult to tune hyper-parameters, we only use $L_\text{CFD}$ for DenseNet201.
\begin{itemize}
    \item 8 to 4-step: $\tau=0.9, \beta=0$.
    \item 4 to 2-step: $\tau=1, \beta=0$.
    \item 2 to 1-step: $\tau=0.85, \beta=0$.
\end{itemize}

\end{document}